\documentclass[runningheads,a4paper]{llncs}
\pdfoutput=1
\usepackage[ruled,linesnumbered,vlined]{algorithm2e}
\usepackage{verbatim}
\usepackage{amsfonts}
\usepackage{epsfig}
\usepackage{algorithmic}
\usepackage{multicol}
\usepackage{multirow}
\usepackage{amsmath}
\usepackage{amssymb}
\usepackage{subfig}
\usepackage{tikz}
\usepackage{pgf}
\usepackage{float}
\usepackage{graphicx}
\usepackage{epstopdf}
\usepackage{comment}
\usepackage{mathptmx}

\usepackage{mathptmx}
\usepackage{geometry}
\usepackage{url}
\urldef{\mailsa}\path|{alfred.hofmann, ursula.barth, ingrid.haas, frank.holzwarth,|
\urldef{\mailsb}\path|anna.kramer, leonie.kunz, christine.reiss, nicole.sator,|
\urldef{\mailsc}\path|erika.siebert-cole, peter.strasser, lncs}@springer.com|    
\newcommand{\keywords}[1]{\par\addvspace\baselineskip
\noindent\keywordname\enspace\ignorespaces#1}

\begin{document}
\title{An ELU Network with Total Variation for Image Denoising}

\author{Tianyang Wang \and Zhengrui Qin \and Michelle Zhu
}

\institute{}

\providecommand{\keywords}[1]{\textbf{\textit{Keywords---}} #1}
\maketitle

\begin{abstract}

In this paper, we propose a novel convolutional neural network (CNN) for image denoising, which uses exponential linear unit (ELU) as the activation function. We investigate the suitability by analyzing ELU's connection with trainable nonlinear reaction diffusion model (TNRD) and residual denoising. On the other hand, batch normalization (BN) is indispensable for residual denoising and convergence purpose. However, direct stacking of BN and ELU degrades the performance of CNN. To mitigate this issue, we design an innovative combination of activation layer and normalization layer to exploit and leverage the ELU network, and discuss the corresponding rationale. Moreover, inspired by the fact that minimizing total variation (TV) can be applied to image denoising, we propose a TV regularized L2 loss to evaluate the training effect during the iterations. Finally, we conduct extensive experiments, showing that our model outperforms some recent and popular approaches on Gaussian denoising with specific or randomized noise levels for both gray and color images. 

\end{abstract}


\section{Introduction}
Image denoising has been a long-time open and challenging research topic in computer vision,  aiming to restore the latent clean image from a noisy observation. Generally, a noisy image can be modeled as $y$ = $x$ + $v$, where $x$ is the latent clean image and $v$ is the additive Gaussian white noise. To restore the clean mapping $x$ from a noisy observation $y$, there are two main categories of methods, namely image prior modeling based and discriminative learning based. Traditional methods, such as BM3D \cite{dabov2007image}, LSSC \cite{mairal2009non}, EPLL \cite{zoran2011learning}, and WNNM \cite{gu2014weighted}, lie in the first category. And the second category, pioneered by Jain et al. \cite{jain2009natural}, includes MLP \cite{burger2012image}, CSF\cite{schmidt2014shrinkage}, DGCRF \cite{vemulapalli2016deep}, NLNet \cite{lefkimmiatis2016non}, and TNRD \cite{chen2017trainable}. Until recently, Zhang et al.~\cite{zhang2017beyond} discovered a deep residual denoising method to learn the noisy mapping with excellent results. However, there is still leeway to boost the denoising performance by reconsidering the activation and the loss function in convolutional neural network (CNN).

In this paper, we propose a deep CNN with exponential linear unit (ELU) \cite{clevert2015fast} as the activation function and total variation (TV) as the regularizer of L2 loss function for image denoising, which achieves noticeable improvement compared to the state-of-the art work \cite{zhang2017beyond} in which the rectified linear unit (ReLU) \cite{krizhevsky2012imagenet} was used as the activation function. By analyzing the traits of ELU and its connection with trainable nonlinear reaction diffusion (TNRD) \cite{chen2017trainable} and residual denoising \cite{zhang2017beyond}, we show that ELU is more suitable for image denoising applications. Specifically, our method is based on residual learning, and the noisy mapping learned with ELU has a higher probability to obtain a desired \textquoteleft energy\textquoteright ~value than that learned with ReLU. It indicates that more noise can be removed from the original noisy observation, hence the denoising performance can be improved. On the other hand, batch normalization (BN) \cite{ioffe2015batch} is also applied in the model for the purpose of training convergence. However, Clevert et al. \cite{clevert2015fast} pointed out that the direct combination of BN and ELU would degrade the network performance. Instead, we construct a new combination of layers by incorporating $1\times$1 convolutional layers, which can better integrate the BN and ELU layers. In our model, we set \textquoteleft Conv-ELU-Conv-BN\textquoteright ~as the fundamental block, where the second \textquoteleft Conv\textquoteright ~denotes the $1\times$1 convolutional layer. Furthermore, we utilize TV, which is a powerful regularizer in traditional denoising methods \cite{chan2005recent,goldluecke2010approach,wang2011mtv}, to regularize L2 loss to further improve the network training performance. Without considering the dual formulation, the TV regularizer can still be solved by stochastic gradient decent (SGD) algorithm during the network training. Finally, we conduct extensive experiments to validate the effectiveness of our proposed approach. 

The main contributions of this work can be generalized in three-folds.  First, we have analyzed the suitability of ELU to denoising task. Second, we have proposed a novel combination of layers to better accommodate ELU and BN. Third, we have applied total variation to regularize L2 loss function. The rest of paper is organized as follows. The proposed network with ELU and TV is presented in section 2 with the analysis of rationale. Extensive experiments and evaluation results can be found in section 3. Section 4 concludes our work with future plan.  
 
\section{The Proposed Network} 
In our approach, a noisy mapping, rather than a clean mapping, is learned since residual learning had been proven successful for image denoising~\cite{zhang2017beyond}. Besides, residual learning had been validated effective for scatter correction in medical image processing \cite{xu2017CT} which requires higher reliability. Before presenting our network architecture, we first discuss the ELU and its intrinsic property for denoising task, followed by how to regularize L2 loss with total variation. Our analysis on both ELU and TV are mainly derived from the energy perspective as denoising is closely relevant to energy reduction. 

\subsection{Exponential Linear Unit}
The primary contribution of an activation function is to incorporate nonlinearity into a stack of linear convolutional layers to increase the network ability of capturing discriminative image features. As one of the activation functions, ELU \cite{clevert2015fast} is defined as:
\begin{equation}
     f(x) =
  \begin{cases}
    x       & \quad \text{if } x > 0 \\
    \alpha (e^x-1)  & \quad \text{if } x \leq 0\\
  \end{cases}
\end{equation}
where parameter $\alpha$ is used to control the level of ELU's saturation for negative inputs and a pre-determined value can be used for the entire training procedure. Unlike ReLU, the most frequently used activation function, ELU does not force the negative input to be zero, which can make the mean unit activation approach zero value since both positive and negative values can counteract each other in the resulted matrix. The near zero mean unit activation not only speeds up learning with a faster convergence but also enhances system robustness to noise. Although ELU has higher time complexity than other activation functions due to the exponential calculation, it can be tolerated if better domain performance is desired.  

\subsection{Motivation of Using ELU} 
\label{Motivation}
For ELU network, Clevert et al. \cite{clevert2015fast} reported a significant improvement on CIFAR-100 classification over the ReLU network with batch normalization. On ImageNet, ELU network also obtained a competitive performance with faster convergence compared to ReLU network. To the best of our knowledge, there is no existing work exploring the connection of ELU with Gaussian image denoising. 
In our work, we note that using different activation functions can generate residual mappings with different \textquoteleft energy\textquoteright, which can be interpreted as angular second moment (ASM) and computed as follows：
\begin{equation}
\label{equation2}
    ASM=\sum_{i,j=0}^{N-1}P_{i,j}^2 
\end{equation}
In practice, $P_{i,j}$ is an element of the gray-level co-occurrence matrix (GLCM) of a noisy mapping: $P_{i,j}$ $\in$ GLCM($v$). Since noisy image has lower ASM compared to a clean one, learning a noisy mapping with lower ASM can be expected. For better clarification, we study the connection between the residual denoising and TNRD \cite{chen2017trainable} which was initially analyzed by Zhang et al. in \cite{zhang2017beyond}. According to their work, such a relation can be described by 
\begin{equation}
\label{equation3}
    v=y-x= \lambda \sum_{k=1}^K (\bar{f_k}*\phi_k(f_k*y))
\end{equation}
where $v$ is the estimated residual of the latent clean image $x$ with respect to the noisy observation $y$. $f_k$ is a convolutional filter used in a typical CNN, and $\bar{f_k}$ is the filter obtained by rotating the filter $f_k$ by 180 degrees. We ignore the constant parameter $\lambda$ since it only weights the right side term in Eq. (\ref{equation3}). The influence function $\phi$ can be an activation function applied to feature maps or the original input. For residual denoising problem, the noisy mapping $v$ should contain as much noise as possible. Therefore, the ASM is expected to be low. According to Eq. (\ref{equation2}), (\ref{equation3}), our goal is to choose the right activation function $\phi$ to have $ASM(v)_{\phi} < ASM(v)_{ReLU}$. To choose an appropriate $\phi$, we conduct a simple experiment on three benchmark datasets, namely Pascal VOC2011, Caltech101, and 400 images of size $180\times$180 from BSD500 dataset that we use to train our network in section \ref{experiments}. 
For each clean image, Gaussian white noise ($\sigma = 25$) is added to obtain the noisy observation denoted by $y$. We generate a randomized $3\times$3 filter as $f_k$, and take ELU as the function $\phi$. The parameter $\alpha$ in Eq. (1) is set to 0.1 for ELU. The comparison of $ASM(v)_{ELU}$ and $ASM(v)_{ReLU}$ is given in Table \ref{table1}.

\begin{table}[]
\vspace{-0.1in}
\centering
\caption{The comparison of $ASM(v)_{ELU}$ and $ASM(v)_{ReLU}$}
\vspace{0.1in}
\label{table1}
\begin{tabular}{|c|c|c|c|}
\hline
                                      & VOC 2011 & Caltech 101 & BSD 400 \\ \hline
$ASM(v)_{ELU} > ASM(v)_{ReLU}$               & 5310     & 3275        & 130     \\ \hline
$ASM(v)_{ELU} < ASM(v)_{ReLU}$               & 9651     & 5868        & 270     \\ \hline
Percentage of $ASM(v)_{ELU} < ASM(v)_{ReLU}$ & 65\%     & 64\%        & 68\%    \\ \hline
\end{tabular}
\vspace{-0.2in}
\end{table}

It can be observed that there is a higher probability to get a lower ASM value when ELU is utilized as the activation function. As mentioned above, a low ASM corresponds to high noisy image. In residual denoising, higher noisy mapping means that more noise can be removed from the original noisy input, resulting in a better denoising effect. In other words, $ASM(v)$ should be small. Therefore, based on Table \ref{table1}, ELU is preferred over ReLU as the activation function for higher noisy residual mapping. 

\subsection{TV Regularizer}  
In Section \ref{Motivation}, we discuss activation selection to reduce ASM energy of a noisy mapping, and we know that the ASM for a noisy image is smaller than that of a clean counterpart. Unlike the ASM, total variation (TV) evaluates the energy directly from the original input signal. A noisy image has larger TV value than that of a clean one, and image denoising can be performed by minimizing the TV value \cite{chan2005recent}. Similarly, in residual denoising, the original L2 loss which measures the distance between the residual mapping and the ground truth noise also needs to be minimized. We thus use TV to regularize L2 loss function which is to be minimized by CNN, and the new loss function is defined as:  
\begin{equation}
\label{equation4}
    L=\frac{1}{2N}\sum_{i=1}^N||R-(y_i-x_i)||^2+\beta TV(y_i-R)
 \end{equation}
 and according to \cite{chan2005recent}, the TV value can be computed by 
 \begin{equation}
    TV(u) \approx \sum_{i,j} \sqrt{(\nabla_x u)^2_{i,j}+(\nabla_y u)^2_{i,j}}
\end{equation}
where we take $R$ as the learned noisy mapping of the latent clean image $x_i$ with respect to the noisy observation $y_i$, and $\nabla_x$, $\nabla_y$ are discretizations of the horizontal and vertical derivatives, respectively. Here, $\{(y_i,x_i )\}_{i=1}^N$ represents the noisy-clean image patch for training. $\beta$ is used to weigh the total variation term. Though $\beta$ can be a fixed value during training, our experiments show that updating its value with the change of training epochs could achieve better results. In general, solving a TV regularizer usually requires the dual formulation, however, it can be solved by stochastic gradient decent (SGD) algorithm during training without considering the dual formulation in our work. In Eq. (\ref{equation4}), the minimization of the first term (L2 loss) will learn the noisy mapping, and the second term (TV) can be regarded as further denoising the obtained clean mapping.

\subsection{Network Architecture}
\label{Architecture}
\begin{figure}
    \centering
    \includegraphics[width=0.8\textwidth]{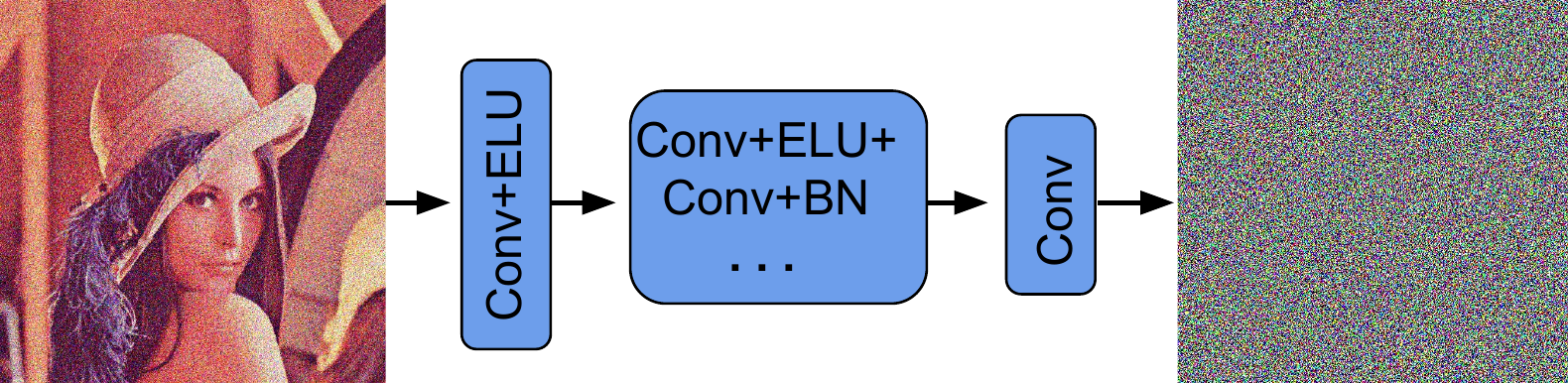}
    \caption{The network architecture with pipe-lined components.}
    \vspace{-0.2in}
    \label{fig1}
\end{figure}

Our model is derived from the vgg-verydeep-19 pre-trained network \cite{simonyan2014very}, and includes a total of 15 convolutional layer blocks and 2 separate convolutional layers. There is no fully connected layer. The network architecture is shown in Fig. \ref{fig1}. 
The first convolutional layer is connected to an ELU layer to add nonlinearity, and the output of the last convolutional layer is fed into the loss layer. Between the two ends, the network is composed of 15 convolutional layer blocks with \textquoteleft Conv-ELU-Conv-BN\textquoteright ~pattern.

It has been shown that ELU can replace ReLU as the activation function in section \ref{Motivation}. Therefore, \textquoteleft Conv-ELU\textquoteright ~is built in each convolutional block. Batch normalization (BN) is necessary for residual denoising as reported by Zhang et al. \cite{zhang2017beyond}. However, direct combination of BN and ELU will adversely affect the network performance \cite{clevert2015fast}. Fortunately, it is known that the pixel-wise co-efficient transformation can be achieved by a $1\times$1 convolutional layer, which can also increase the non-linearity of the decision function \cite{simonyan2014very,he2016deep}. We thus utilize a $1\times$1 convolutional layer between ELU and BN layer. Every second \textquoteleft Conv\textquoteright ~in each block holds $1\times$1 filters, and other filters are all in the size of $3\times$3. Such configuration not only exerts the advantages of $1\times$1 convolutional layer, but also avoids direct connection of BN and ELU.  

Note that our model does not contain any pooling layer since the final output must have the same size as the original input. One may argue that fully convolutional networks (FCN) \cite{long2015fully} can also restore the output size, however it cannot be used in our case because it contains a pooling layer and thus needs up-sampling operation, which is not desirable for image denoising. Furthermore, FCN was originally designed for pixel-level classification without fully considering the relationships between pixels.

\section{Experiments}
\label{experiments}
Our experiments are conducted in Matlab using MatConvNet framework \cite{vedaldi2015matconvnet}, which provides convenient interface to design network structure by adding or removing predefined layers. One NVidia Geforce TITAN X GPU is used to accelerate the mini-batch processing. To validate the efficacy of our method, we train three networks. The first network is for gray image Gaussian denoising with specific noise levels; the second and the third one are for color image Gaussian denoising with specific and randomized noise levels, respectively.

\subsection{Data sets}
We choose the experiment datasets similar to the work from~\cite{zhang2017beyond}. For gray image denoising with a specific noise level, 400 images of size $180\times$180 from Berkeley segmentation dataset (BSD500) are used for training and $128\times$1600 patches are cropped with size $40\times$40 for each. All color images are converted to gray ones prior to training. Three noise levels are considered, namely $\sigma =15, 25, 50$. Two testing datasets are used: BSD68 that contains 68 images,  and the other set of 12 most frequently used gray images\footnote{https://github.com/cszn/DnCNN/tree/master/testsets/Set12} in image processing community. Note that there is no overlapping between the training and the testing datasets.

For color image denoising, the color version of BSD68 is employed as the testing data and the remaining 432 images from BSD500 are used for training. $\sigma =15, 25, 50$ are still used as the specific noise levels, and $128\times$3000 patches with size $50\times$50 are cropped. However, for blind denoising, the noise levels are randomly selected from range [0, 55]. 

\subsection{Compared Methods}
Besides the well-known methods such as BM3D\cite{dabov2007image}, LSSC\cite{mairal2009non},  WNNM\cite{gu2014weighted}, EPLL\cite{zoran2011learning}, MLP\cite{burger2012image}, CSF\cite{schmidt2014shrinkage}, we also consider another four similar neural network based methods, namely DGCRF\cite{vemulapalli2016deep}, NLNet\cite{lefkimmiatis2016non}, TNRD\cite{chen2017trainable} and DnCNN\cite{zhang2017beyond}, since these methods have reported promising results.  

\subsection{Network Training}
As explained in Section \ref{Architecture}, our network has 15 convolutional blocks and 2 separate convolutional layers. We use the same depth for both gray and color image denoising. We initialize the weights using MSRA as He et al.~\cite{he2015delving} did for image classification. The TV regularizer is incorporated into the L2 loss function, and the entire network is trained by SGD with a momentum of 0.9. The initial learning rate is set to be 0.001, and changed to 0.0001 after 30 out of 50 epochs. The initial value of $\beta$ in Eq. (\ref{equation4}) is set to 0.0001, and increased to 0.0005 after 30 epochs. The weight decay is set to 0.0001. It is worth noting that weight decay regularizes the filter weights, whereas total variation regularizes the L2 loss.  

\subsection{Results Analysis}
In our work, peak signal-to-noise ratio (PSNR) is utilized to evaluate the denoising effect. We first compare our method with other well-known methods on BSD68 gray images. The results are given in Table \ref{table2}, where the best ones are highlighted in bold. It can be seen that our model shows the best average PSNR for all the three specific noise levels. When $\sigma =50$, our method outperforms BM3D by 0.7dB, which reaches the estimated upper bound over BM3D in \cite{dabov2007image}. We further validate our method on the 12 commonly used test images for image processing task, and the average PSNR is compared in Table \ref{table3}. Our method outperforms DnCNN by around 0.1dB, which gives similar increments as in Table~\ref{table2}.

\begin{table}[!ht]
\vspace{-0.1in}
\centering
\caption{The average PSNR of different methods on the gray version of BSD68 dataset.}
\vspace{0.1in}
\label{table2}
\begin{tabular}{|c|c|c|c|c|c|c|c|c|c|c|c|}
\hline
\multicolumn{1}{|l|}{Methods} & \multicolumn{1}{l|}{BM3D} & \multicolumn{1}{l|}{MLP} & \multicolumn{1}{l|}{EPLL} & \multicolumn{1}{l|}{LSSC} & \multicolumn{1}{l|}{CSF} & \multicolumn{1}{l|}{WNNM} & \multicolumn{1}{l|}{DGCRF} & \multicolumn{1}{l|}{TNRD} & \multicolumn{1}{l|}{NLNet} & \multicolumn{1}{l|}{DnCNN} & \multicolumn{1}{l|}{Ours} \\ \hline
$\sigma$ = 15                        & 31.08                     & -                        & 31.21                     & 31.27                     & 31.24                    & 31.37                     & 31.43                      & 31.42                     & 31.52                      & 31.73                        & \textbf{31.82}            \\ \hline
$\sigma$ = 25                        & 28.57                     & 28.96                    & 28.68                     & 28.71                     & 28.74                    & 28.83                     & 28.89                      & 28.92                     & 29.03                      & 29.23                        & \textbf{29.34}            \\ \hline
$\sigma$ = 50                        & 25.62                     & 26.03                    & 25.67                     & 25.72                     & -                        & 25.87                     & -                          & 25.96                     & 26.07                      & 26.23                        & \textbf{26.32}            \\ \hline
\end{tabular}
\end{table}

\begin{table}[!ht]
\centering
\caption{The average PSNR of different methods on the 12 most commonly used gray images in image processing community.}
\vspace{0.1in}
\label{table3}
\begin{tabular}{|c|c|c|c|c|c|c|c|c|}
\hline
Methods        & BM3D  & WNNM  & EPLL  & MLP   & CSF   & TNRD  & DnCNN & Ours           \\ \hline
$\sigma$  = 15 & 32.37 & 32.70 & 32.14 & -     & 32.32 & 32.50 & 32.86 & \textbf{32.96} \\ \hline
$\sigma$ = 25  & 29.97 & 30.26 & 29.69 & 30.03 & 29.84 & 30.06 & 30.44 & \textbf{30.55} \\ \hline
$\sigma$ = 50  & 26.72 & 27.05 & 26.47 & 26.78 & -     & 26.81 & 27.21 & \textbf{27.29} \\ \hline
\end{tabular}
\end{table}

Besides gray image denoising, we also train our model with specific and randomized noise levels for color image denoising. Table \ref{table4} depicts the competency of our model trained with specific noise levels. Similar to gray image case, our method increases the PSNR by about 0.1dB compared to DnCNN, which is trained with specific noise levels as well. Note that training with randomized noise levels also generates satisfied results, which, however, are inferior to the results achieved by the models trained with specific noise levels. 

\begin{table}[!ht]
\centering
\caption{The average PSNR of different methods on the color version of BSD68 dataset.}
\vspace{0.1in}
\label{table4}
\begin{tabular}{|c|c|c|c|c|c|c|}
\hline
Methods       & CBM3D & MLP   & TNRD  & NLNet & DnCNN & Ours           \\ \hline
$\sigma$ = 15 & 33.50 & -     & 31.37 & 33.69 & 33.99   & \textbf{34.10} \\ \hline
$\sigma$ = 25 & 30.69 & 28.92 & 28.88 & 30.96 & 31.31   & \textbf{31.41} \\ \hline
$\sigma$ = 50 & 27.37 & 26.01 & 25.96 & 27.64 & 28.01   & \textbf{28.11} \\ \hline
\end{tabular}
\vspace{-0.2in}
\end{table}

The visual comparison between our method and other well-known methods are given in Fig. 2 $\sim$ Fig. 4. We add noise ($\sigma =25$) for one gray image, and our model is trained with a specific noise level ($\sigma =25$). The denoising effect is shown in Fig. 2. While in Fig. 3 and Fig. 4, to validate randomized level (blind) denoising effect, we add two different noise levels ($\sigma = 35,50$) for each color image, respectively. Note that color denoising visual comparison is carried only between our method and DnCNN, since DnCNN, to our best knowledge, is the state-of-the-art denoising method. Moreover, DnCNN also supports blind denoising. We compare our model with the version of DnCNN which was trained with randomized noise levels in the range of [0,55]. To achieve fair comparison, our model is also trained with randomized noise levels within the same range. Results show that our model preserves more image details. Moreover, the over-smooth issue of the background scene is also alleviated. Hence, the utilization of total variation does not over-smooth the image.  

\begin{figure*}[!hbt]
  \vspace{-0.1in}
  \centering
  \subfloat[Clean]
  {\includegraphics[width=0.2435\textwidth]{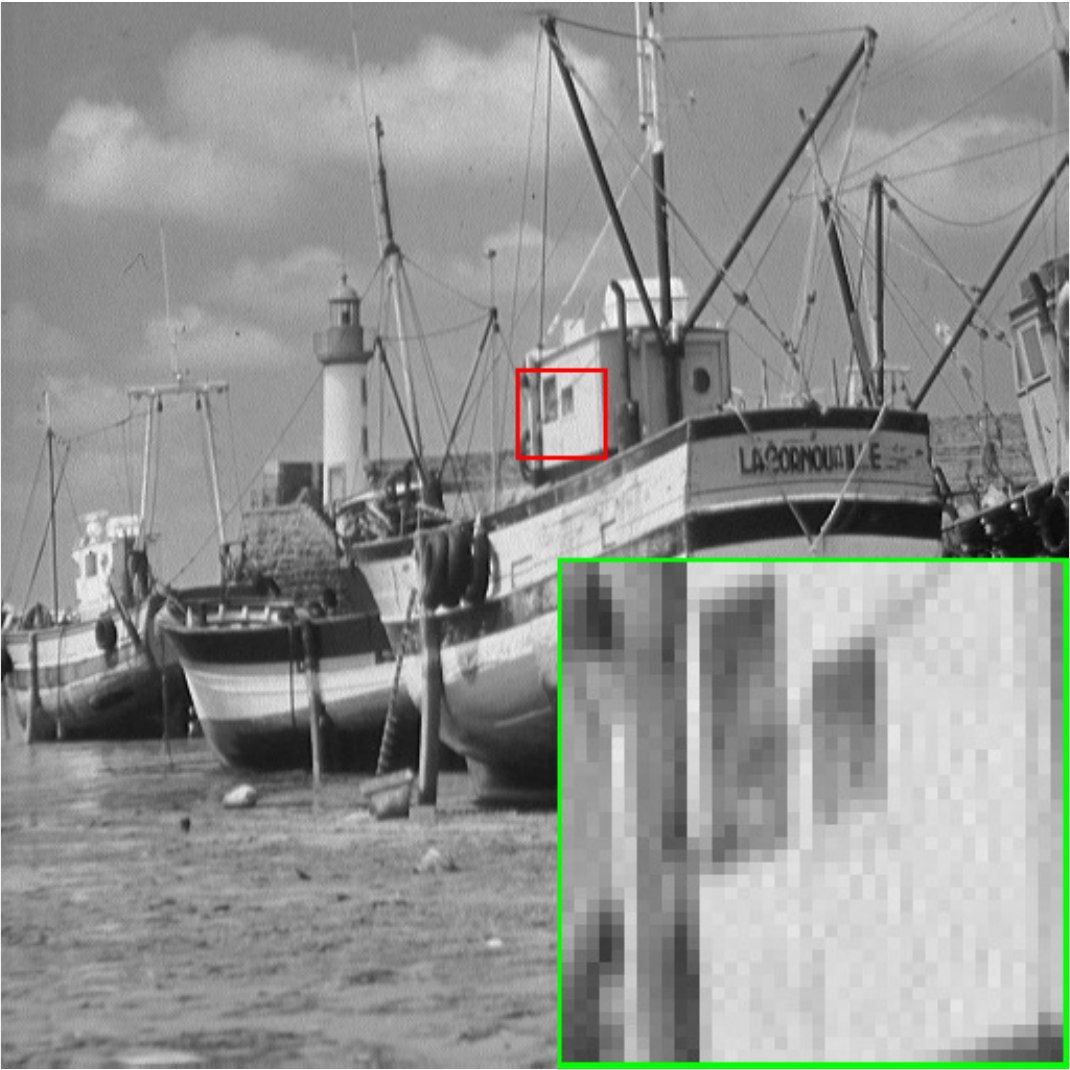}
  \label{fig2a}}
  \subfloat[Noisy/20.18dB]
  {\includegraphics[width=0.2435\textwidth]{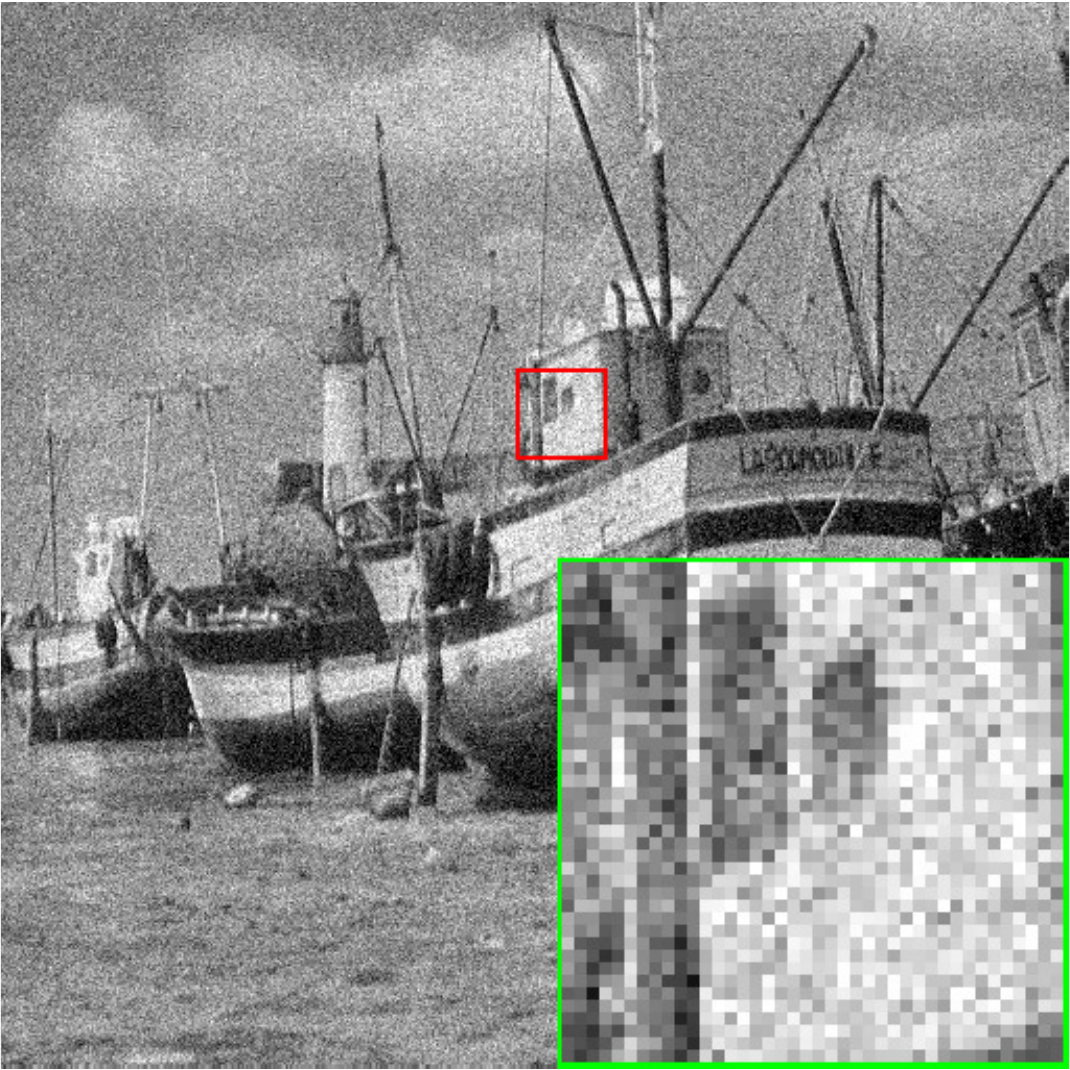}
  \label{fig2b}}
  \hfill
  \subfloat[BM3D/29.91dB]
  {\includegraphics[width=0.2435\textwidth]{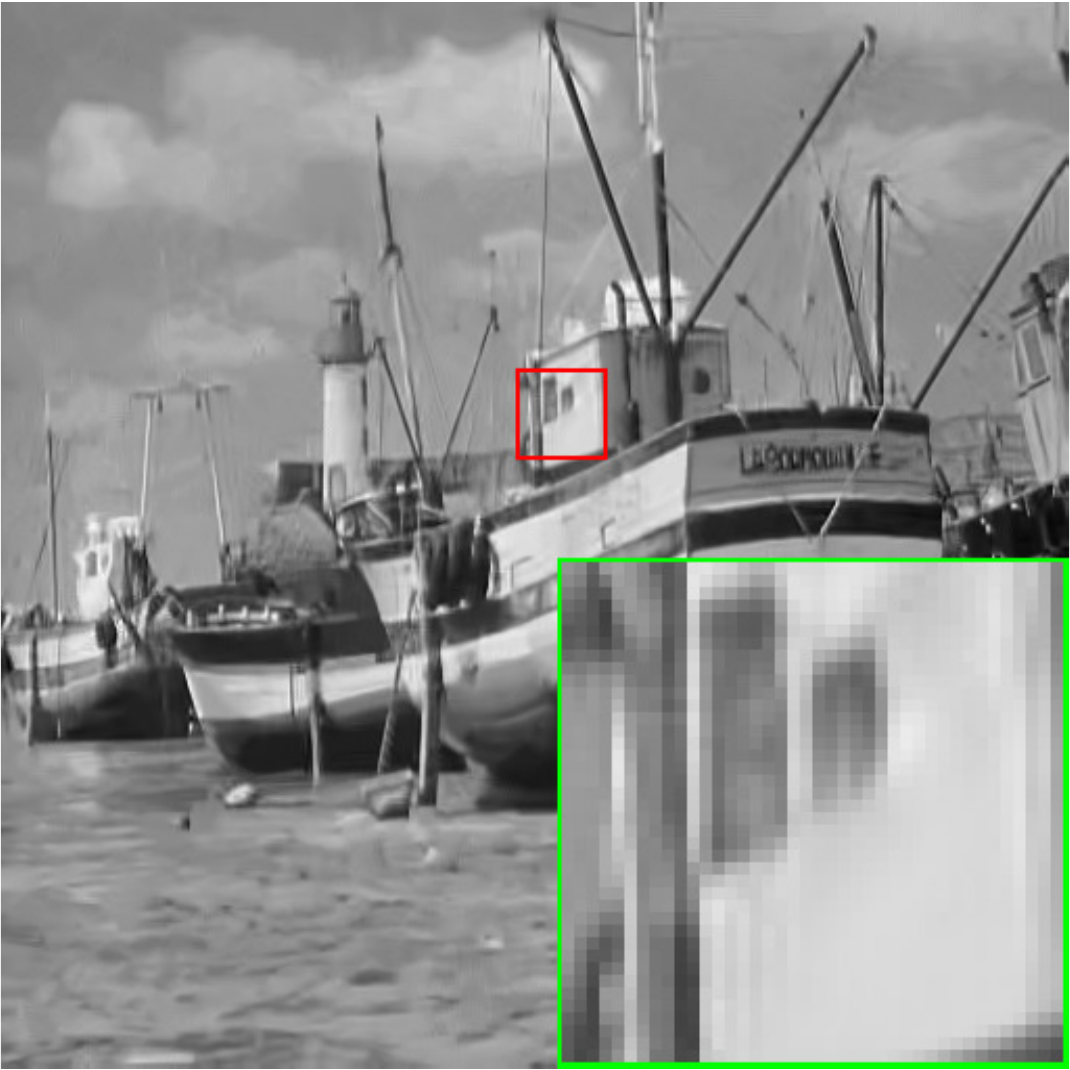}
  \label{fig2c}}
    \subfloat[MLP/29.95dB]
  {\includegraphics[width=0.2435\textwidth]{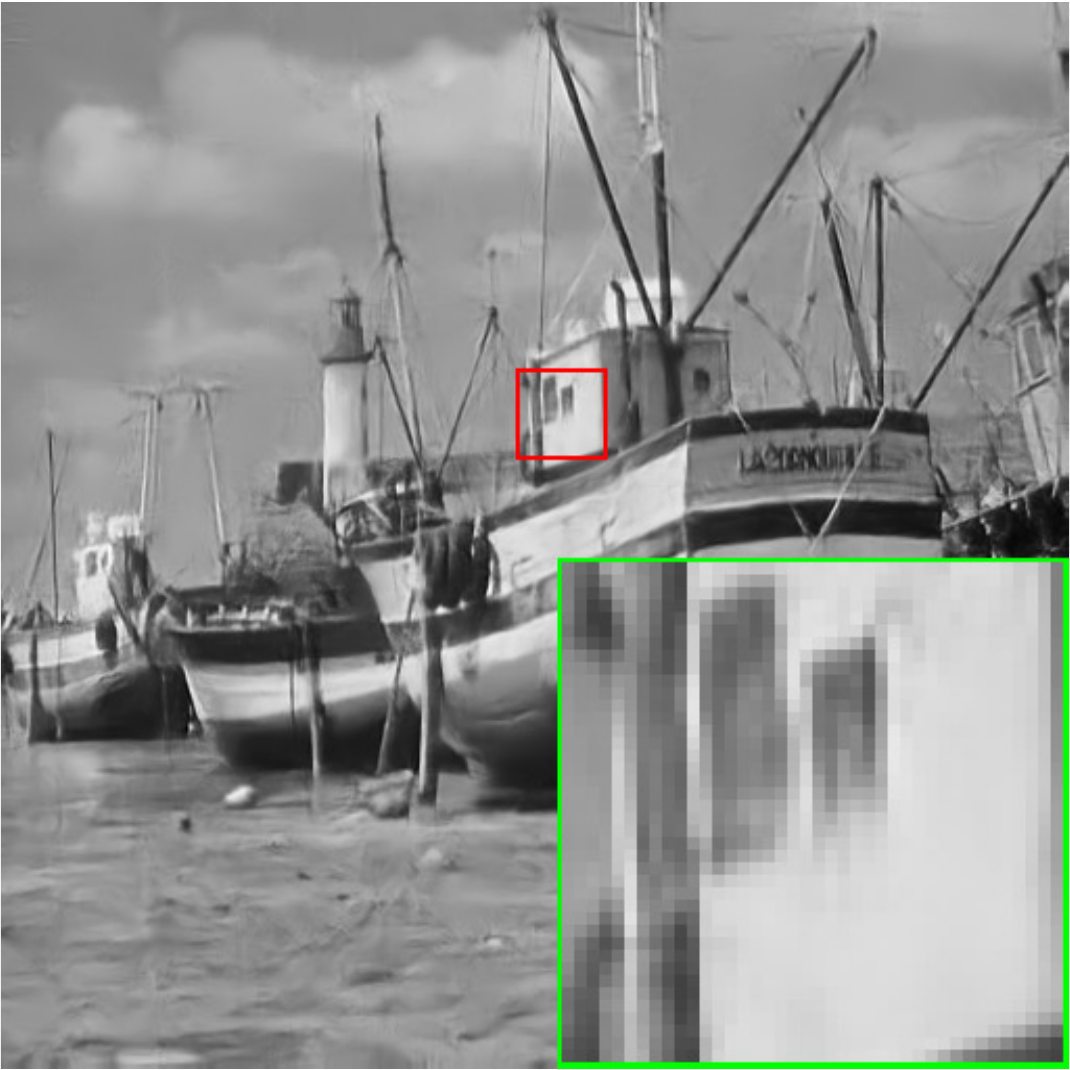}
  \label{fig2d}}
  
  \subfloat[TNRD/29.92dB]
  {\includegraphics[width=0.2435\textwidth]{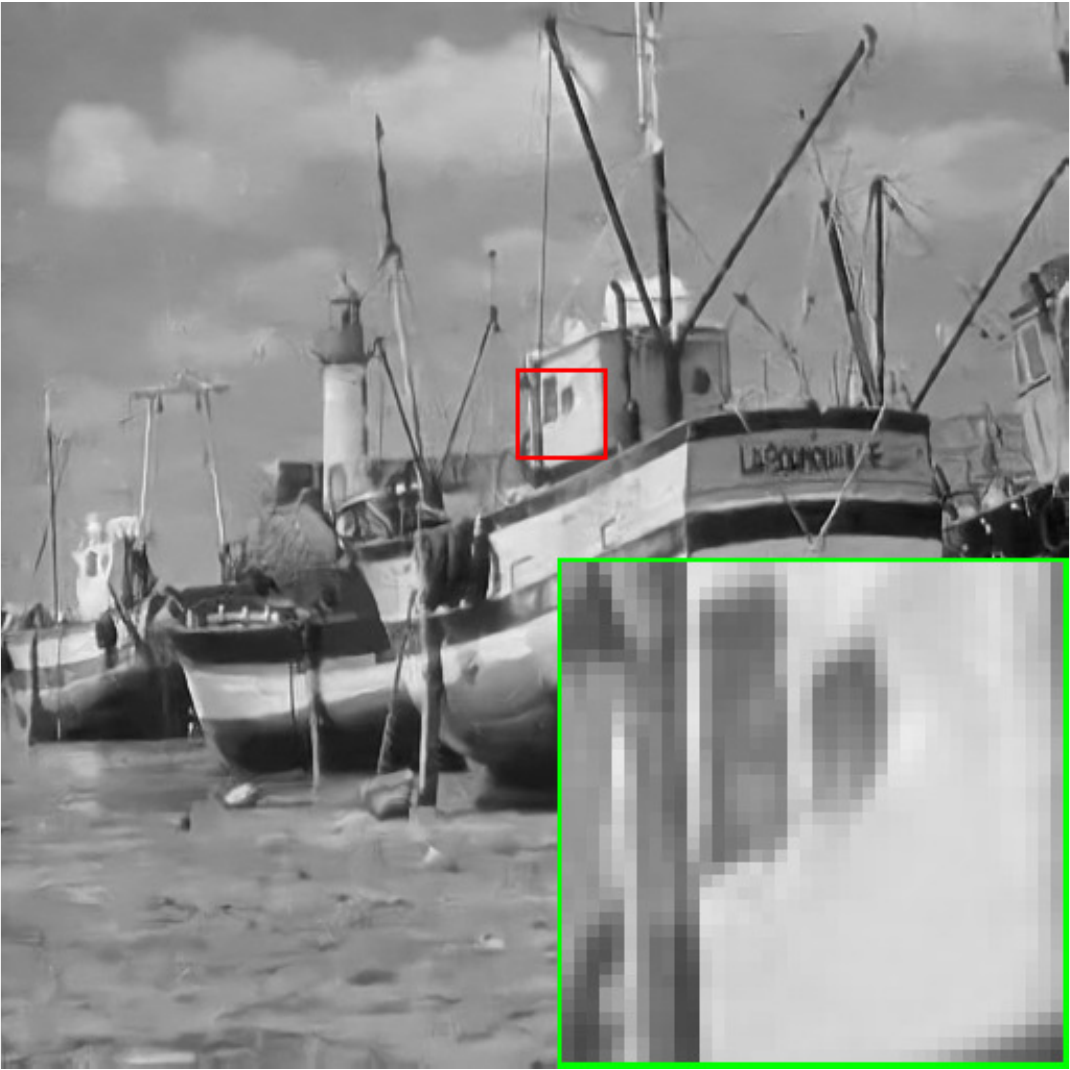}
  \label{fig2e}}
  \hfill
  \subfloat[WNNM/30.03dB]
  {\includegraphics[width=0.2435\textwidth]{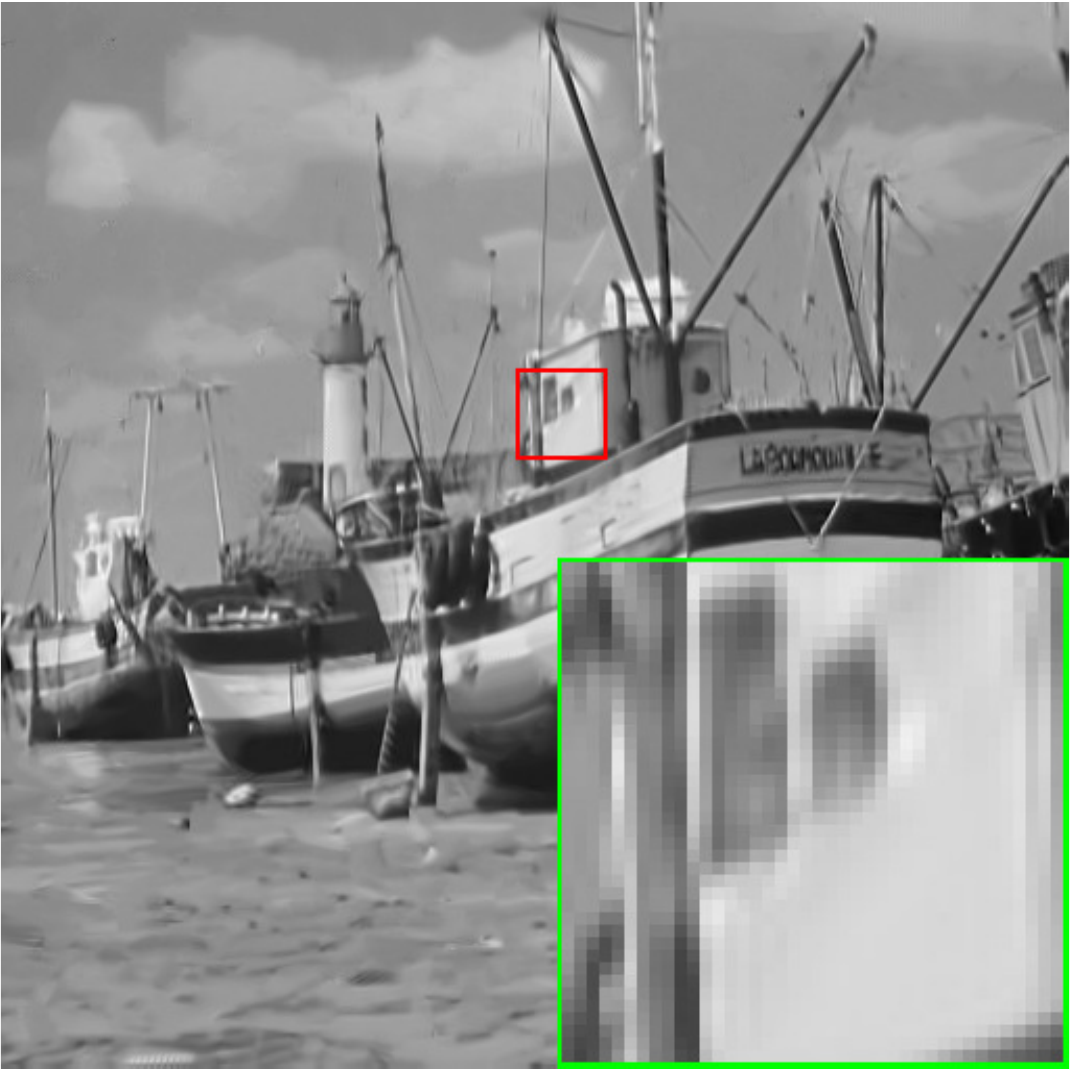}
  \label{fig2f}}
    \subfloat[DnCNN/30.22dB]
  {\includegraphics[width=0.2435\textwidth]{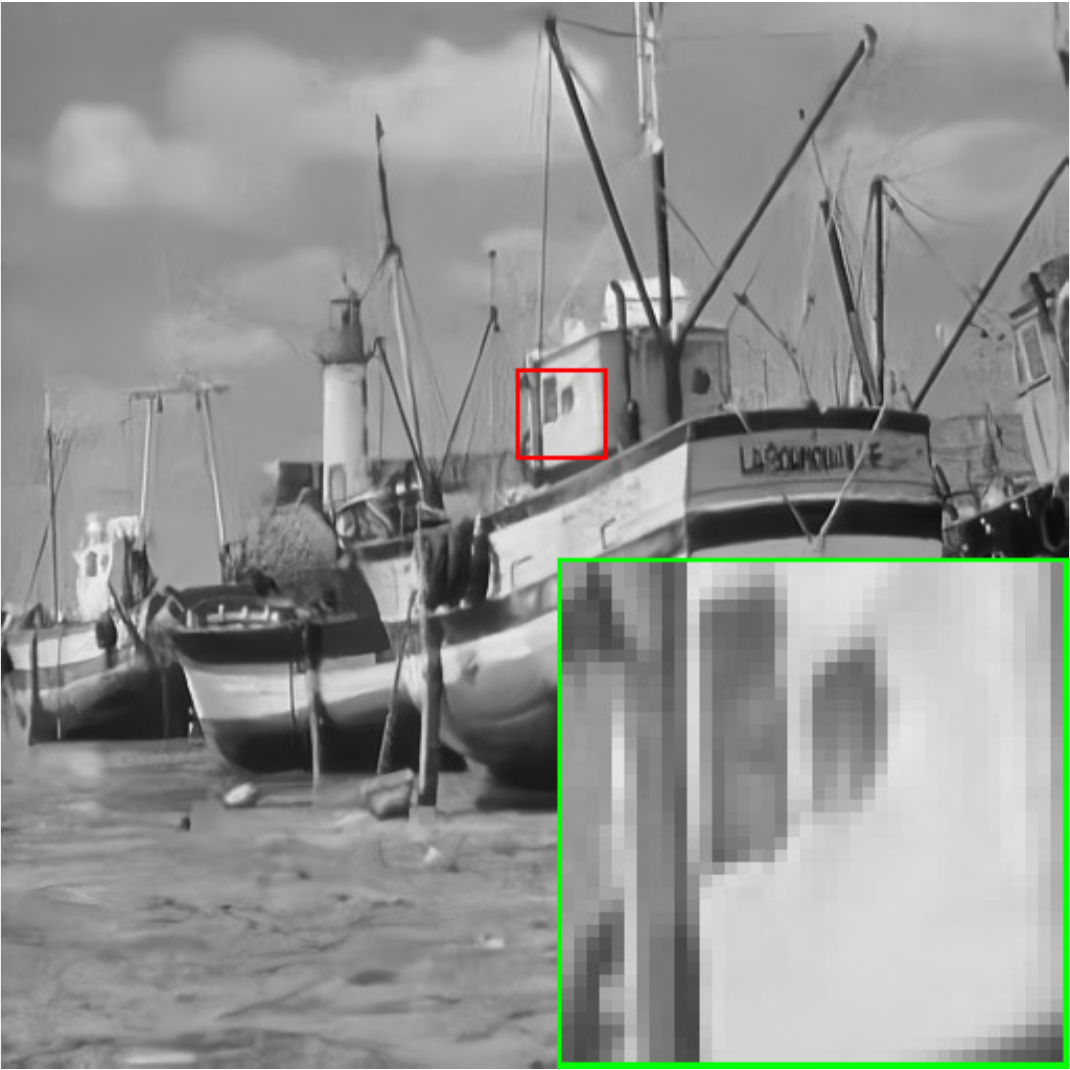}
  \label{fig2g}}
    \subfloat[Ours/30.32dB]
  {\includegraphics[width=0.2435\textwidth]{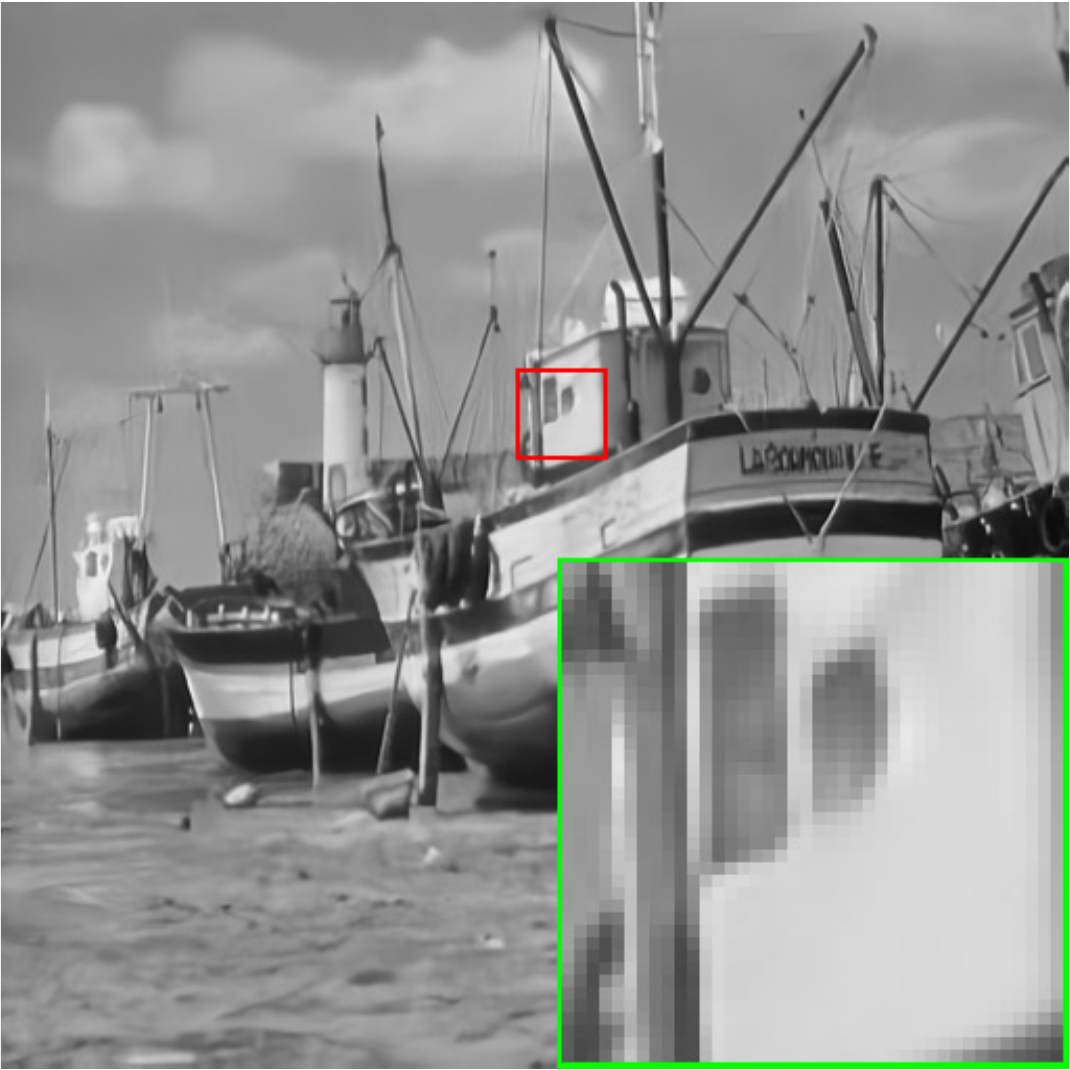}
  \label{fig2h}}

  \caption{Visual comparison of gray image denoising between our method and other methods. Our model is trained with specific noise level ($\sigma=25$). The clean image is polluted by noise ($\sigma=25$).}
  \vspace{-0.2in}
  \label{figure2}
\end{figure*}

\begin{figure*}[!hbt]
  \centering
  \subfloat[Clean]
  {\includegraphics[width=0.2435\textwidth]{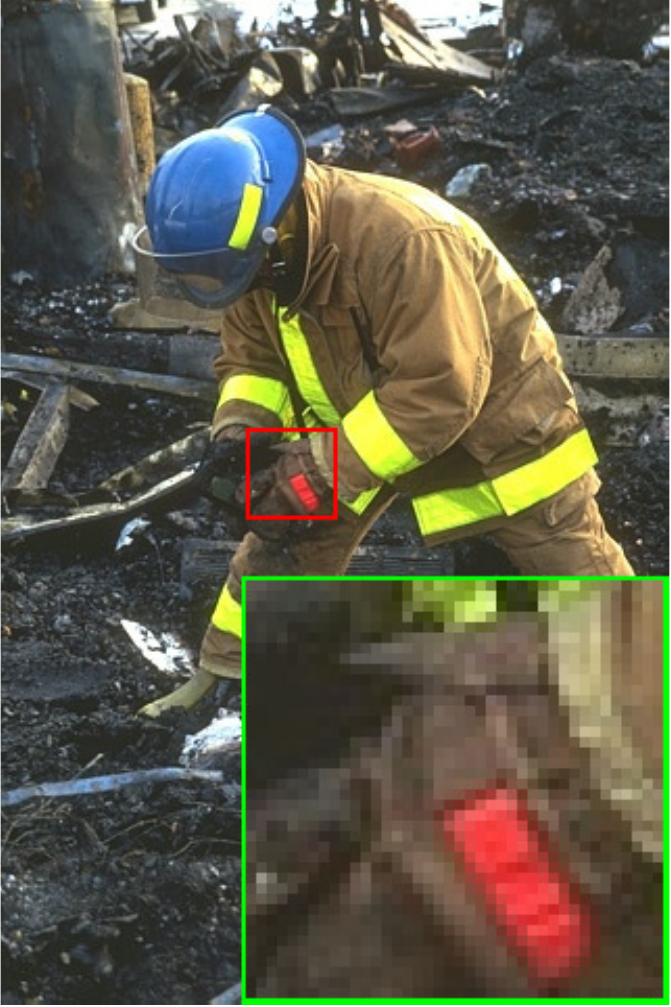}
  \label{fig3a}}
  \hfill
  \subfloat[Noisy/17.70dB]
  {\includegraphics[width=0.2435\textwidth]{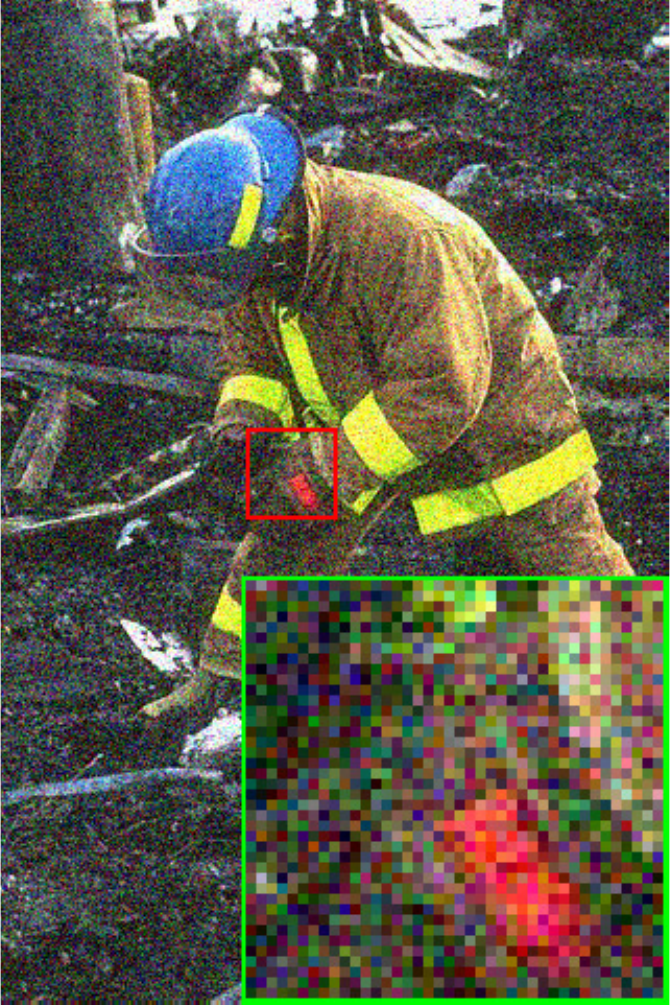}
  \label{fig3b}}
    \subfloat[DnCNN/28.18dB]
  {\includegraphics[width=0.2435\textwidth]{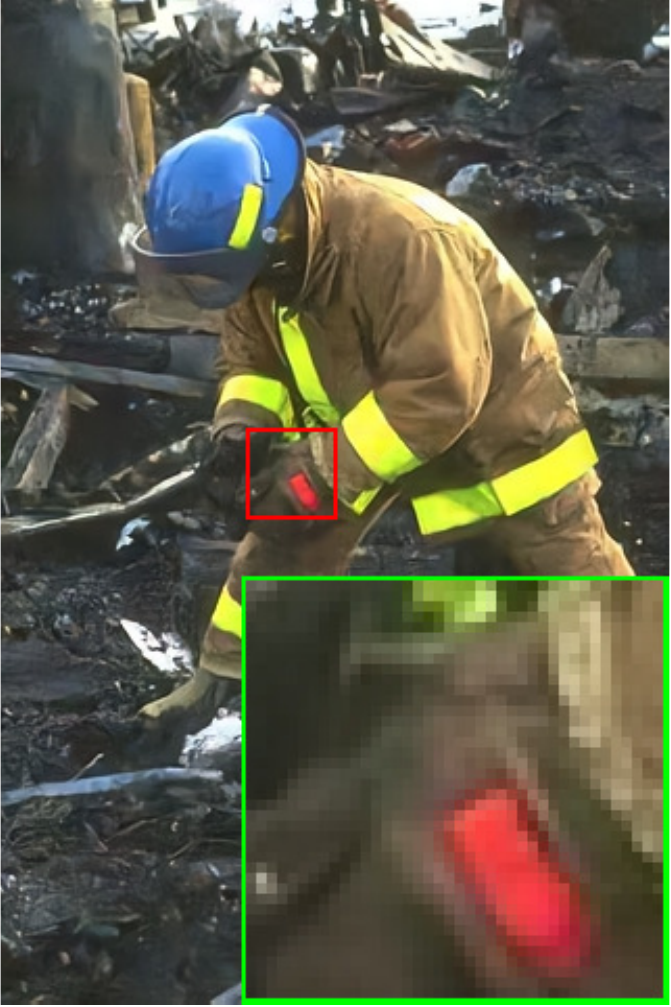}
  \label{fig3c}}
    \subfloat[Ours/28.28dB]
  {\includegraphics[width=0.2435\textwidth]{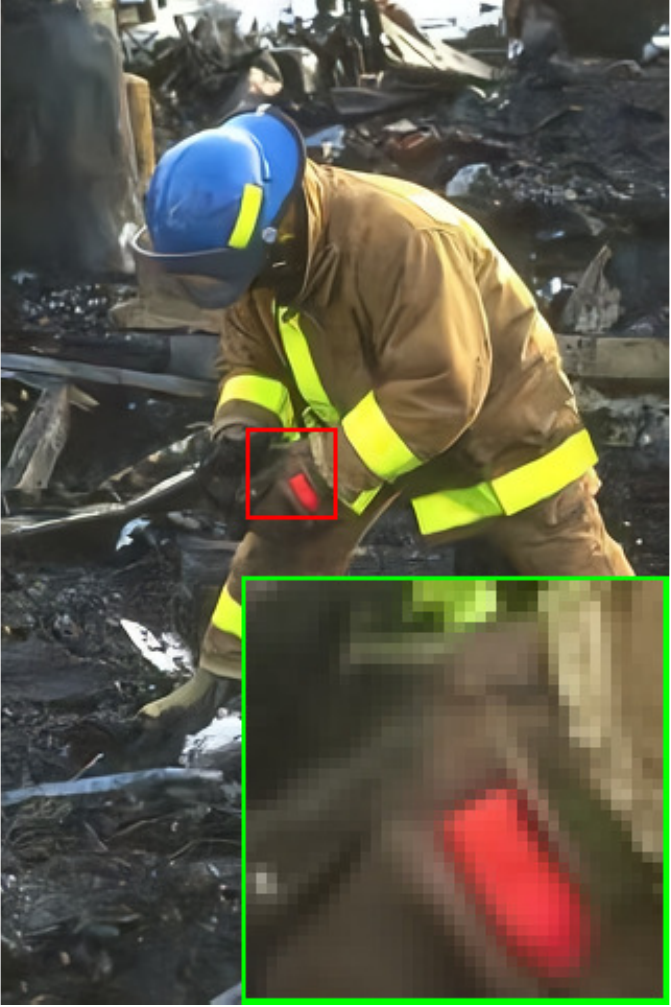}
  \label{fig3d}}
  
  \caption{Visual comparison of color image denoising between our method and DnCNN. Both models are trained with randomized noise level from range [0,55]. The noise ($\sigma=35$) is added to the clean image.}
  \vspace{-0.2in}
  \label{figure3}
\end{figure*}

\begin{figure*}[!hbt]
  \vspace{-0.1in}
  \centering
  \subfloat[Clean]
  {\includegraphics[width=0.2435\textwidth]{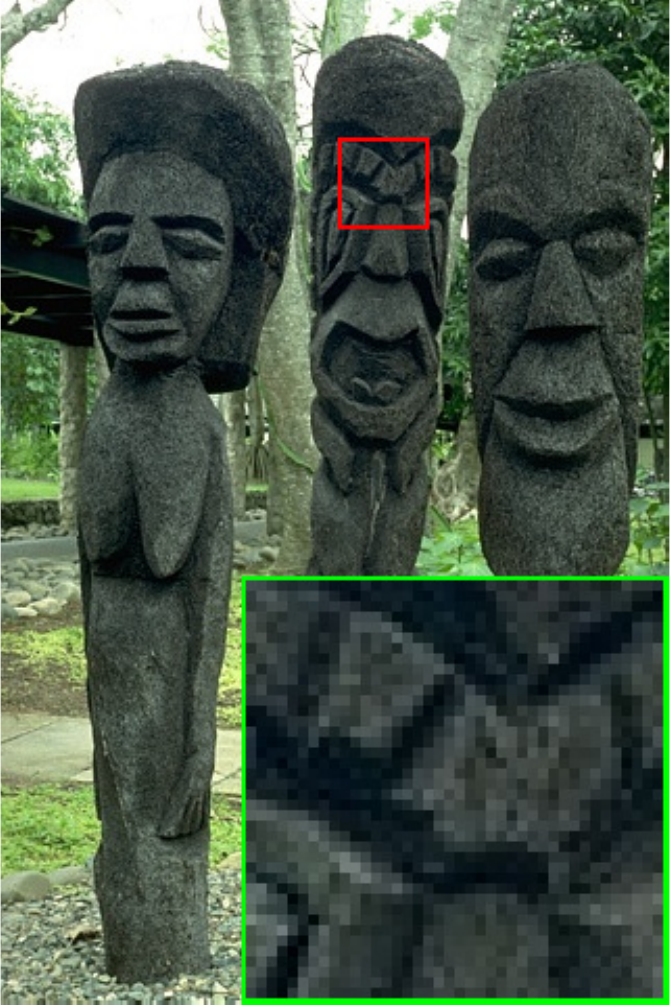}
  \label{fig4a}}
  \hfill
  \subfloat[Noisy/15.10dB]
  {\includegraphics[width=0.2435\textwidth]{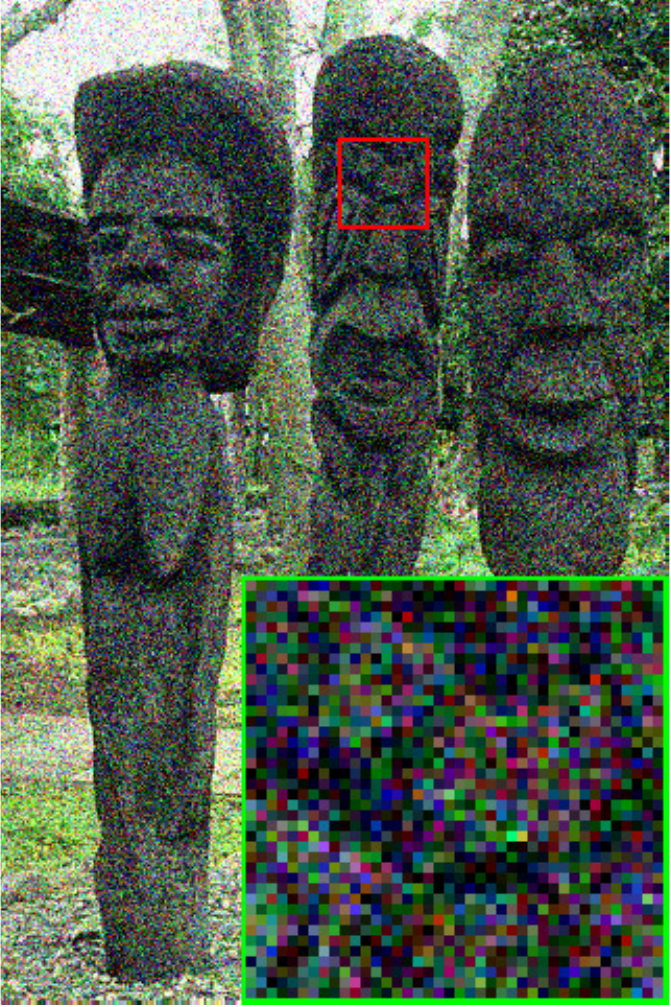}
  \label{fig4b}}
    \subfloat[DnCNN/24.97dB]
  {\includegraphics[width=0.2435\textwidth]{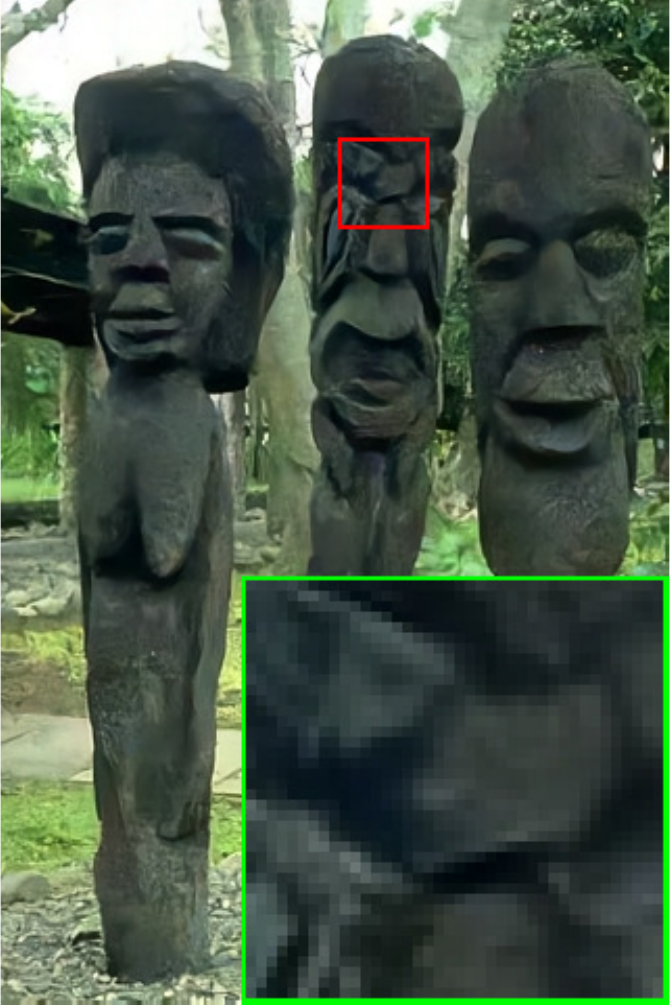}
  \label{fig4c}}
    \subfloat[Ours/25.06dB]
  {\includegraphics[width=0.2435\textwidth]{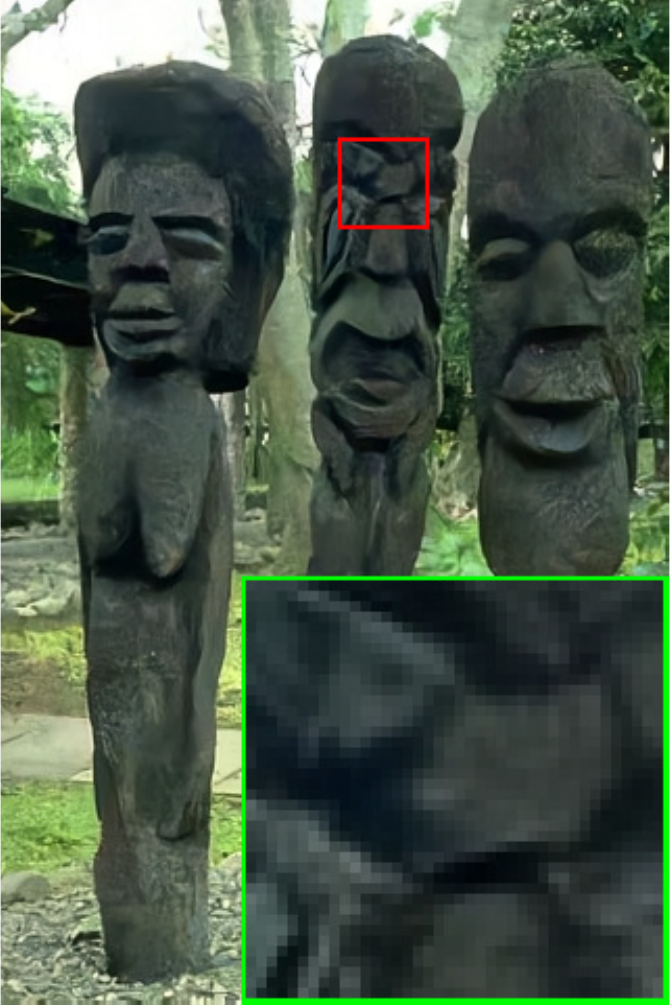}
  \label{fig4d}}
  
  \caption{Visual comparison of color image denoising between our method and DnCNN. Both models are trained with randomized noise level from range [0,55]. The noise ($\sigma=50$) is added to the clean image.}
  \label{figure4}
\end{figure*}

\section{Conclusion}
In this paper, we propose a deep convolutional neural network with exponential linear unit as the activation function and total variation as the regularizer of L2 loss for Gaussian image denoising. By analyzing the advantages of ELU and the connection with residual denoising and trainable nonlinear reaction diffusion model, we have validated that ELU is more suitable for image denoising problem. To better accommodate ELU and BN layer, we design a novel structure by incorporating $1\times$1 convolutional layer. By studying the traits of total variation, we have shown the feasibility of regularizing L2 loss with TV in convolutional nets. Extensive experiments show promising quantitative and visual results compared with other reputed denoising methods which are based on image prior modeling or discriminative learning.    
Furthermore, since we have observed the improvement in image denoising by replacing ReLU with ELU in convolutional nets, we will continue to investigate other improved versions of ReLU, such as PReLU, RReLU, and LReLU for denoising task in future work. 

\section*{Acknowledgments}\label{sec:Acknowledgments}
This project was partially supported by the new faculty start-up research grant at Montclair State University.

\bibliographystyle{splncs03}
\bibliography{Reference.bib} 

\end{document}